\documentclass[10pt, a4paper]{article}

\usepackage{lrec-coling2024} % this is the new style

\usepackage{xspace}

\newcommand{\eg}{\textit{e}.\textit{g}.~} 
\newcommand{\wrt}{\textit{w}.\textit{r}.\textit{t}}

\def\model{ACT-MNMT\xspace}
\def\hardmodel{TECT-MNMT\xspace}
\usepackage{threeparttable}
\usepackage{multirow}
\usepackage{color}
\usepackage{booktabs}
\usepackage{subfigure}

\title{ACT-MNMT: Auto-Constriction  Turning for Multilingual Neural Machine Translation}

\name{Shaojie Dai$^{1,2,3}$, Xin Liu$^{2}$, Ping Luo$^{1,2,3}$, Yue Yu$^{2}$}

\address{$^{1}$Key Laboratory of Intelligent Information Processing\\
Institute of Computing Technology, Chinese Academy of Sciences\\
$^{2}$Peng Cheng Laboratory, $^{3}$University of Chinese Academy of Sciences \\
         daishaojie22@mails.ucas.ac.cn, hit.liuxin@gmail.com, luop@ict.ac.cn, yuy@pcl.ac.cn\\}

\abstract{
Large language model (LLM) has achieved promising performance in multilingual machine translation tasks through zero/few-shot prompts or prompt-tuning. However, due to the mixture of multilingual data during the pre-training of LLM, the LLM-based translation models face the off-target issue in both prompt-based methods, including a series of phenomena, namely instruction misunderstanding, translation with wrong language and over-generation. For this issue, this paper introduces an \textbf{\underline{A}}uto-\textbf{\underline{C}}onstriction \textbf{\underline{T}}urning mechanism for \textbf{\underline{M}}ultilingual \textbf{\underline{N}}eural \textbf{\underline{M}}achine \textbf{\underline{T}}ranslation (\model), which is a novel supervised fine-tuning mechanism and orthogonal to the traditional prompt-based methods.
In this method, \model automatically constructs a constrained template in the target side by adding trigger tokens ahead of the ground truth.
Furthermore, trigger tokens can be arranged and combined freely to represent different task semantics, and they can be iteratively updated to maximize the label likelihood. Experiments are performed on WMT test sets with multiple metrics, and the experimental results demonstrate that \model achieves substantially improved performance across multiple translation directions and reduce the off-target phenomena in the translation. % We release our code anonymously in: \url{https://anonymous.4open.science/r/ACT-MNMT-43E6/}.
 \\ \newline  }

\begin{document}

\maketitleabstract

\section{Introduction}
\label{introduction}
% encoder-decoder 和 decoder的不同，以及为什么选择en-de(在Transformer打开了大语言模型的理论窗口之后，大语言模型发展出了三种路线。第一种，以Google BERT、ELECTRA为代表的Encoder-Only（编码器）路线；第二种，以Google T5、BART为代表的Encoder- Decoder（编解码器）路线；第三种，以OpenAI GPT为代表的Decoder-Only（解码器）路线。
% 这三种路线，Encoder-Only路线适合理解类任务，很难应对生成式任务，也不具有好的扩展性和适应性，虽然被Google BERT在个别子领域一度带火，但现在几乎处于被主流抛弃的地步。Encoder- Decoder路线适合特定场景任务，但通用性和扩展性也比较差。Decoder-Only路线首先非常适合生成类任务，同时对各类任务都具有很好的通用性，在工程上也具有很高的可扩展性（scale），非常适合将模型规模做大。 https://mp.weixin.qq.com/s/1EPPOqSpGyoF5koBeeEuVg)

% 多语优点：多语LLM不是在平行语料上训练的
% 多语缺点：
% 传统缺点: 会存在训练多个模型的问题，全量微调会损失模型的通用信息
% 相关工作可重点讲解ICL，其次off-target-ratio

% % trigger token的优点：1.隐式地描述任务 2.相比hard constraint 占用了更少的token，信息更密集，减少推理时间 3.传统token的任务描述token数量不定 4.任务描述的方式不止一种很难找到最佳的 5.比上下文学习的性能更好，微调数据也很少
% 根据：AutoPrompt: Eliciting Knowledge from Language Models with Automatically Generated Prompts
% 固定的template对简单的关系会有较好的捕获，但对抽象的关系难以捕获
% 从模型中本身抽取 和平行语料中训练的连个分布是不同的
Large language models (LLMs) have demonstrated remarkable capabilities in various scenarios, displaying a level of aptitude that approaches or even outperforms human-level intelligence in certain aspects.
Among these diverse competencies of LLMs, the multilingual neural machine translation (MNMT) ability of LLMs has gained significant attention from academia and industry. Some researches~\cite{vilar2022prompting, hendy2023good, bawden2023investigating} have investigated and presented a comprehensive evaluation of numerous LLMs for multilingual neural machine translation performance, such as the performance of different models in comparison with state-of-the-art baselines.

\begin{figure}[h]
	\centering 
% 	\vspace{-6mm}
% 	\hspace{-5mm} 
	\subfigure[BLOOMZ-7B1]{
	    \label{fig_bloomz_otr}
		\includegraphics[width=.51\linewidth]{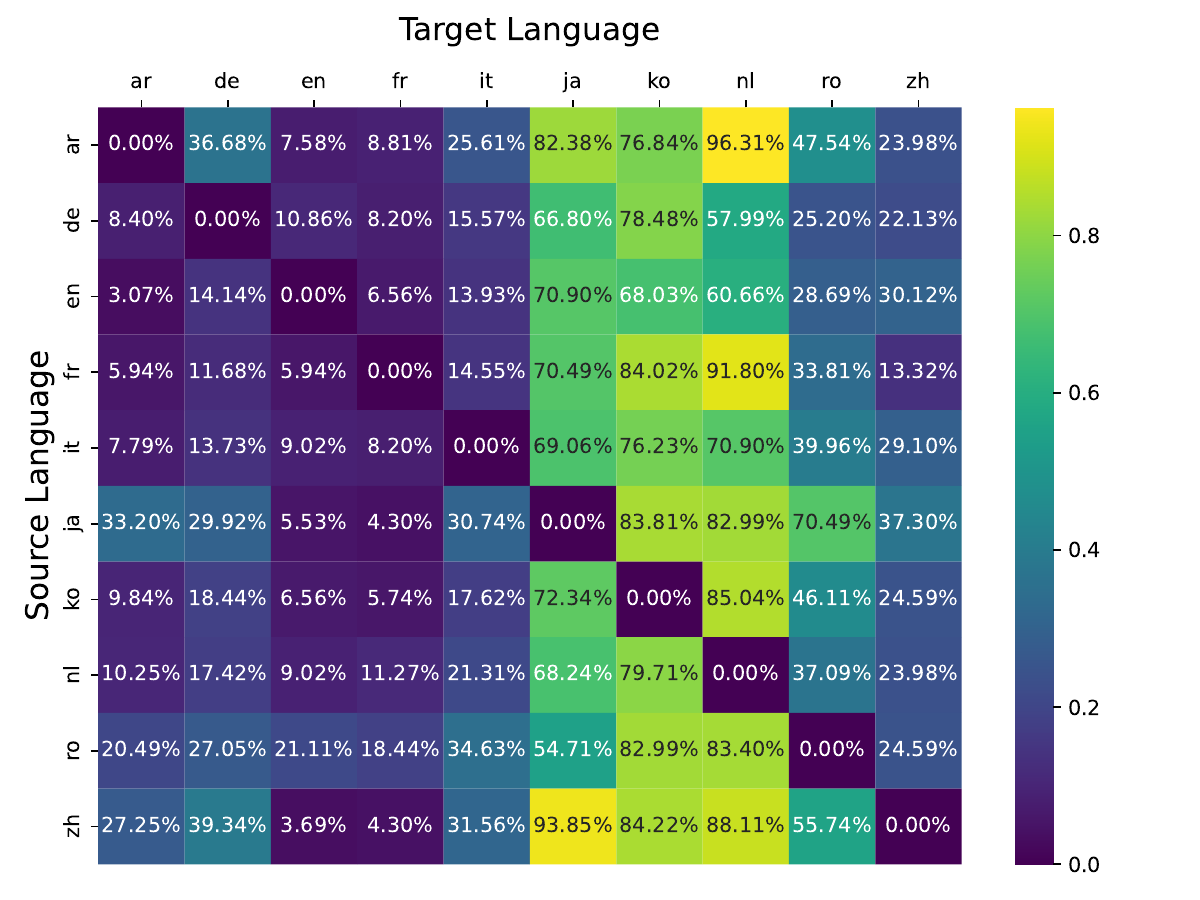} 
	} 
% 	\quad{}  
	\hspace{-9mm}
	\subfigure[mT0-xl (3.7B)]{
	    \label{fig_mt0_otr}
		\includegraphics[width=.51\linewidth]{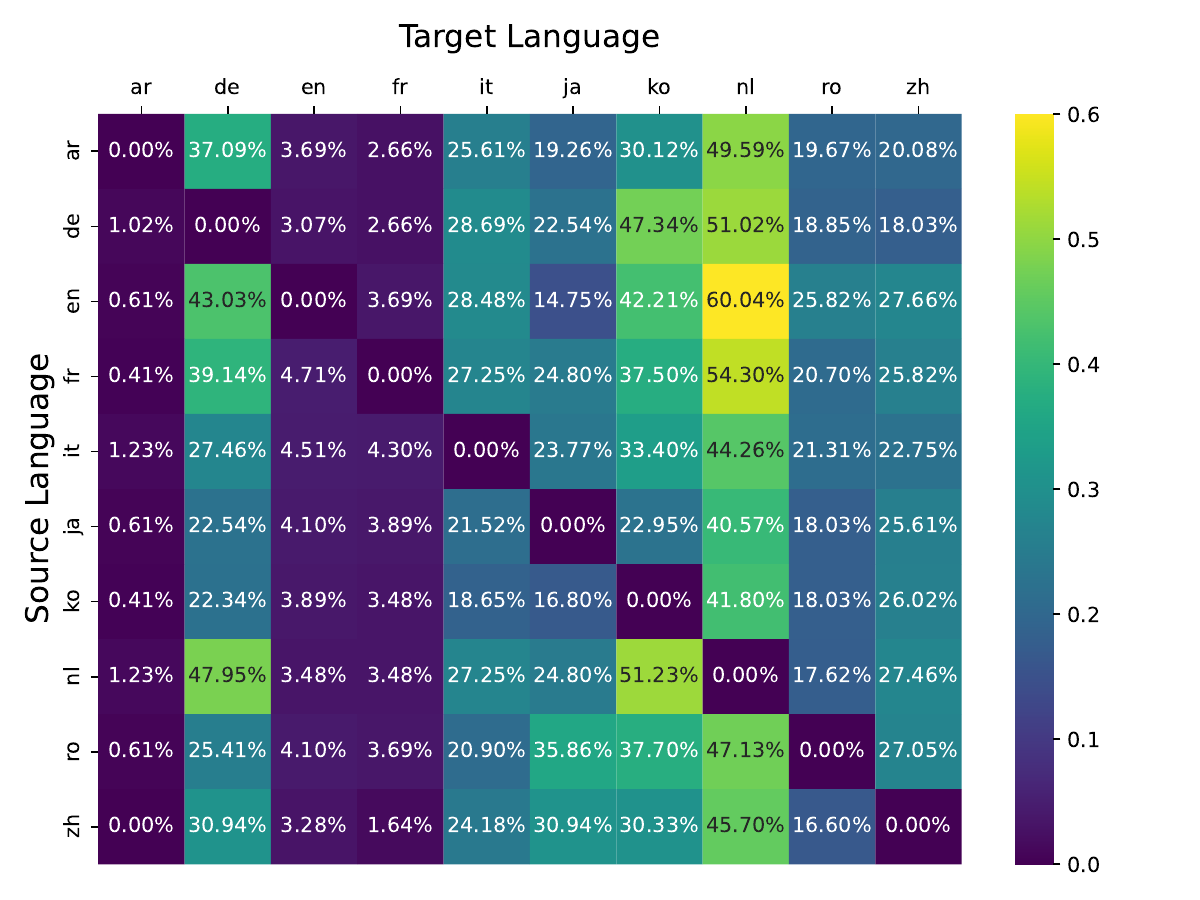}
	} 
% 	\vspace{-2mm}
	\caption{Off-target ratio on IWSLT 2017 test datasets (to evaluate the off-target ratio between any pair of languages, we extract a set of 488 sentences from the IWSLT 2017~\cite{cettolo2017overview} test dataset that have identical English content, and each sentence has 10 different translations in various languages).}
	\label{fig_otr}
% 	\vspace{-3mm}
\end{figure}

However, as reported by~\citet{bawden2023investigating}, in the BLOOM model, MNMT zero-shot prompt seriously suffers from the off-target issue of over-generation and generating in the wrong language. This phenomenon not only appears in generative pre-training (GPT) style models, but also in encoder-decoder style models. Figure~\ref{fig_otr} shows the off-target ratio of BLOOMZ-7B1 and mT0-xl (3.7B) models on IWSLT 2017 test sets. In Figure~\ref{fig_otr}, we utilize the \texttt{langdetect} package\footnote{https://github.com/Mimino666/langdetect} to detect the highest probability language of the generated translations.
As shown in the Figure \ref{fig_otr}, we have the following observations: 1) both BLOOMZ-7B1 and mT0-xl~\cite{muennighoff2022crosslingual} exhibit a serious off-target issue under the zero-shot setting with the prompt of "{\textit{Translate the following text from \{sourceLanguage\} to \{targetLanguage\}: \{sourceString\}}}" used in the multitask fine-turning stage.
2) Low-resource languages present more serious off-target issue, such as Japanese (ja), Korean (ko), and Dutch (nl), especially between non-English language pairs.
3) mT0-xl shows a better translation performance compared to BLOOMZ-7B1, even though the later has more parameters.
%We argue that the encoder-decoder style model is more suitable for the conditional generation task.
We also show the over/under generation results on IWSLT 2017 in section \ref{auxiliary_experiments}, Figure \ref{fig_oug}.
We exploit the over/under-generation ratio as an indicator of LLM's comprehension on instructions, where a lower ratio denotes a stronger ability to understand the instructions.

\citet{vilar2022prompting} have investigated various strategies for choosing translation demonstrations for few-shot prompting to probe the MNMT ability of PaLM. The research concludes that demonstrations quality is the most important factor, and when there are no demonstrations, it seriously affects the quality of translation. 
This is in line with human intuition.
Furthermore, eliciting and probing the translation ability of LLMs still face four major challenges in off-target issue:
%(using Figure \ref{fig_introduction} as an example to illustrate the following challenges). 
1) \textit{misunderstanding of instructions (MI)}: when we feed a LLM with a translation-related instruction and source sentence in English, the expected gold output should be the target sentence in Chinese. But the LLM may misunderstand the instruction and treat it as a classification-related instruction, leading to an incorrect output.
2) \textit{off-target translation (OT)}: the LLM outputs a sentence that is not in the target language.
3) \textit{source copy (SC)}: LLM outputs a sentence that is almost identical to the source sentence, which can also be considered as a type of OT error.
4) \textit{over/under generation (OUG)}: LLM translations are significantly longer or shorter than references.

% To address these challenges, to explore the translation ability of LLMs and enhance their understanding of instructions, 
To address these challenges and explore the translation abilities of LLMs while enhancing their understanding of instructions,
in this work, we focus on automatically constructing constrained template in the target side to assist LLMs to generate the desired outputs.
First, we propose two types of trigger tokens: common trigger tokens and specific trigger tokens. The common trigger tokens are designed to capture shared information across different translation directions. The specific trigger tokens aim to capture target language-specific information.
Second, we design a mapping strategy to establish an association between trigger token sequences and different translation directions.
Trigger tokens can be considered as a continuous latent representation of different translation directions.
Overall, our primary contributions can be summarized as follows:
\begin{itemize}

    \item We propose a novel task-enhanced constrained turning method (\hardmodel), which utilizes a manually designed hard encoding constrained template to assist LLMs in understanding the prompt and guiding the model's output.
    
    \item We further extent \hardmodel to auto-constriction turning (\model), which uses a soft encoding constrained template that do not require manual design to constrain the model output.% and extracts LLMs constraint as additional supervised information.

    % \item We conduct extensive experiments on WMT 22 dataset. Results show that both \hardmodel and \model could effectively alleviate four types of translation errors mentioned in Figure \ref{fig_introduction} and outperforms instruction fine-tuning baseline.
    \item We conduct extensive experiments on WMT dataset. Results show that both \hardmodel and \model can effectively alleviate four types of translation errors mentioned above and outperforms instruction fine-tuning baseline.

    % The experimental results demonstrate that, \model achieves substantially improved performance across a variety of translation directions, and even competitive with much larger models.
    
\end{itemize}
% On the other hand, compared to AutoPrompt needs search the optimal trigger tokens from the whole vocabulary, our method only need to search in the optimal trigger tokens pool (contains trigger tokens for different tasks, does not intersect with the vocabulary token set, and the number is much smaller than the number of vocabulary).

% Firstly, \textit{sensitive to this constrains}, constrains unfortunately requires manually crafting the context to feed into the model. This process is both time-consuming and unintuitive for various tasks. Furthermore, it is crucial to acknowledge that models are sensitive to this constrains context, and inadequately constructed contexts will result in diminished performance. Simplifying the requirement to manually define constrains would enhance the utility of constrains as an analytical instrument.

% To enable these services and keep improving the quality, deep learning model architecture evolves rapidly, and the model size is also growing at a tremendous speed. For example, from GPT to GPT-3 the model size increased 1500x in 2 years.

\section{Related Work}
% part0: hard soft prompt
% part1: prompt[Prompting PaLM for Translation: Assessing Strategies and Performance] 
% part2: off-target issue
% part3: MNMT
% part4: constraint generation
% part5: peft
% part6: LLM
% Large language models (LLMs) have emerged as a fundamental aspect of natural language processing due to their ability to significantly enhance data efficiency on relevant tasks \cite{liu2022few}. 

% Using LLMs for multilingual machine translation is attracting more and more attention.
LLMs have acquired advanced language comprehension skills during the pre-training stage, and shown remarkable abilities in multilingual translations~\cite{lin2022few}.
The traditional method to probe the translation performance of LLM is In-Context Learning (ICL)~\cite{brown2020language}, which is powerful and can provide a few demonstrations to guide models on how to perform, even on an unseen task without fine-tuning. Researchers have been devoted to investigating and designing demonstrations selection strategy. \citet{zhang2023prompting} find that prompt template and demonstration selection both have substantial impact on translation.
\citet{agrawal2022context} shows that the translation quality and the domain of the in-context examples matter greatly.

However, LLMs with a parameter size below 7 billion exhibit a weak capacity for ICL, which may underestimate their translation abilities~\cite{li2023eliciting}. Recently, AutoPrompt~\cite{shin2020autoprompt} employs discrete text prompts to maximize the log-likelihood of true labels. While this technique outperforms manually designed prompts, there is still a gap compared to model tuning, and discrete prompts are usually difficult for humans to understand. 
Therefore, P-Tuning~\cite{liu2021p}, Prompt Tuning~\cite{lester2021power} and Multitask Prompt Tuning~\cite{wang2023multitask} have been developed to enhance the LLM's ability in downstream tasks through optimize prompts in a continuous way.
Recently, some researchers attempt to strike a balance between the costs of fine-tuning, inference, and the translation performance of ICL, such as mFTI~\cite{li2023eliciting}, which collects a small portion of high-quality multilingual parallel sentences (1k per language pair) to fine-tune XGLM-7B \cite{lin2022few}, and then explores the multilingual translation performance.
However, all the methods mentioned above focus on making efforts on the input side of the model.

\section{Problem}
% \begin{Def} [Multilingual Neural Machine Translation] is an important task in the field of natural language processing (NLP), which aims to translate a sentence $x$ in source language into a sentence $y$ in target language \cite{huang2022unifying, costa2022no}. Formally, given a set of language pairs $L$, the MNMT model is trained on the combination of $|L|$ parallel corpus $D=\{D_{l}^{train}\}_{l=1}^{|L|}$, where $D_{l}^{train}$ denotes the train dataset of language pair $(X_l, Y_l)$. The MNMT model is commonly trained to minimize the following negative log-likelihood loss:

\textbf{Multilingual Neural Machine Translation} is an important task in the field of natural language processing (NLP), which aims to translate a sentence $x$ in source language into a sentence $y$ in target language~\cite{huang2022unifying, costa2022no}. Formally, given a set of language pairs $L$, the MNMT model is trained on the combination of $|L|$ parallel corpus $D=\{D_{l}^{train}\}_{l=1}^{|L|}$, where $D_{l}^{train}$ denotes the training data of language pair $(X_l, Y_l)$. The MNMT model is commonly trained to minimize the negative log-likelihood loss as follows:

\begin{equation}
\mathcal{L}=\sum_{D_l \in D} \sum_{(x,y) \in D_l} \sum_{i \leq |y|} -\log p(y_{i} | x,y_{<i}; \theta),
\end{equation}
where $p(\cdot | \cdot; \theta)$ maps the source sentence and previous generated text $y_{<i}$ to the next target token, and $\theta$ denotes the model's parameters.
\begin{figure*}[t]
\centering
\includegraphics[width=1.0\textwidth]{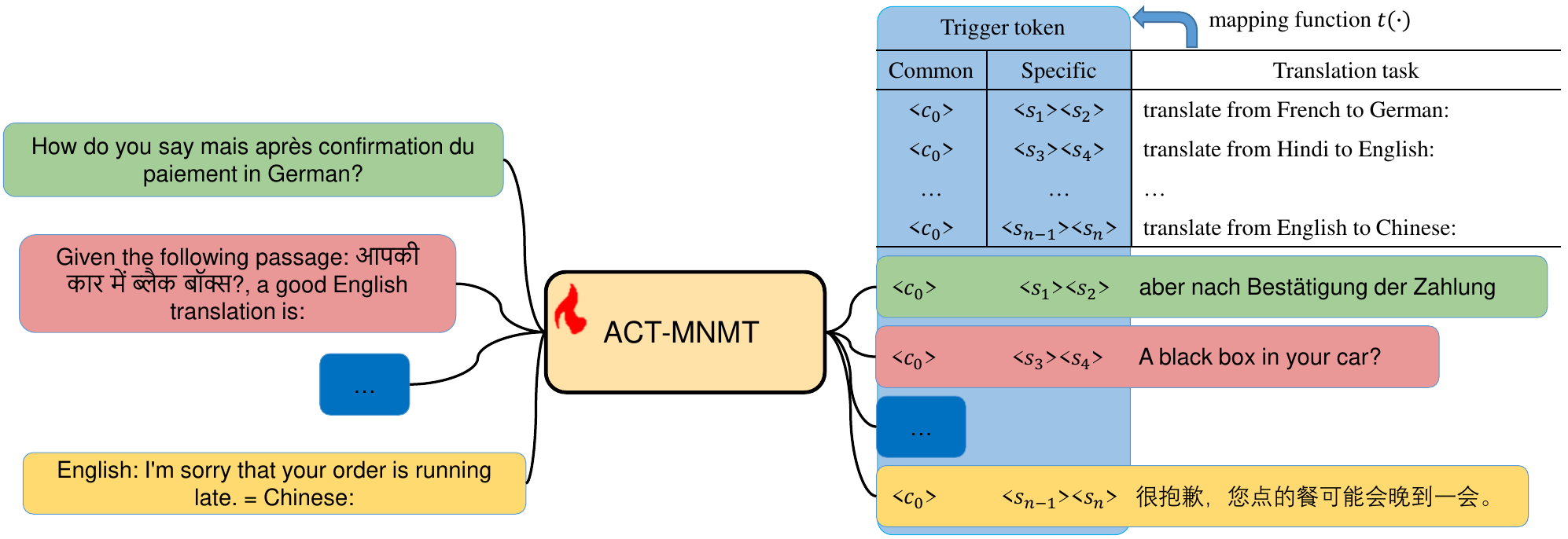} % Reduce the figure size so that it is slightly narrower than the column.
\caption{An overview of \model applied to Multilingual Neural Machine Translation (the number of common trigger token for each translation direction is 1, and the number of specific trigger token is 2 in this example).}
\label{fig_framwork}
\end{figure*}

We should note that depending on different MNMT model training strategies, the source sentence $x$ and target sentence $y$ may be combined with other information (\eg, joining the prompt and source sentence together as the model input).
In the following part of this paper, We will continue to use $x$ to denote the model input and $y$ to denote the model output.

% \begin{Def} [Auto-Constrained Turning] first generates the trigger tokes related to prompts, then generate the ground truth $y$:
% \begin{equation}
% \mathcal{L}=\sum_{(x,y) \sim D} \sum_{i \leq |z|} -\log p(z_{i} | x,z_{<i}; \theta),
% \end{equation}
% where $z=[t(x), y]$, $[\cdot,\cdot]$ denotes the contact operation, and $t(\cdot)$ represents a mapping function for auto-constrained generation (\eg, in the upper right part of Figure \ref{fig_introduction}, $t(\cdot)$ can map the translation task to a series of trigger tokens). Trigger tokens are special tokens added to the vocabulary during fineturning , which can also be seen as additional supervised information to determine whether the model understands the instructions.
% % 它也可以被看作是额外的监督信息，判断模型是否理解了指令
% \end{Def}
% We should note that trigger tokens are generated in the target side, therefore, the added trigger tokens can be arbitrarily combined and arranged, allowing them to represent even unseen tasks semantic information during the fine-tuning stage. As a result, auto-constrained turning performs well under zero-shot settings.% 由于trigger token是在生成端的，所以添加的trigger token 能够任意组合排列，即使是在为微调阶段没有见过的任务也可能通过其表示出来，因此在zero-shot上性能较好

\section{Methods}

Intuitively, a natural way to constrain and guide LLM to generate the optimal output is: first, generating the descriptive information related to the task in the prompt, and then generating the output, which is similar to Chain-of-Thought~\cite{wei2022chain}.
Therefore, we can determine whether LLM understand the prompts based on this auxiliary descriptive information.
However, writing descriptive information is time-consuming and it is also unclear whether the same phrasing will be effective for every model.
To this end, we first propose a task-enhanced constrained turning mechanism for multilingual neural machine translation (\hardmodel), which utilises a hard encoding constrained template.
Then, we extend \hardmodel to an auto-constriction turning mechanism (ACT-MNMT), which utilises a soft encoding constrained template. The illustration of \model is provided in figure~\ref{fig_framwork}.

\begin{table*}[!ht]

\caption{MT prompts for instruction turning. All prompts specify the target language (L2, in blue and in brackets). Six prompts are also specified the source language (L1, in red and in brackets) in the instruction.}

\label{table_instruction}
\begin{center}
\begin{tabular}{ll}
\toprule
Prompt    & Target  \\
\midrule
Given the following source text in \textcolor{red}{[L1]}: [SRC], a good \textcolor{blue}{[L2]}     & \multirow{2}{*}{translate from \textcolor{red}{[L1]} to \textcolor{blue}{[L2]}: [TGT]} \\
translation is: & \\

If the original version says [SRC] then the \textcolor{blue}{[L2]} version should say:   &  translate from \textcolor{red}{[L1]} to \textcolor{blue}{[L2]}: [TGT] \\

What is the \textcolor{blue}{[L2]} translation of the sentence: [SRC]?   & translate from \textcolor{red}{[L1]} to \textcolor{blue}{[L2]}: [TGT] \\

\textcolor{red}{[L1]}: [SRC] = \textcolor{blue}{[L2]}:   &  translate from \textcolor{red}{[L1]} to \textcolor{blue}{[L2]}: [TGT] \\

[SRC] translates into \textcolor{blue}{[L2]} as:  & translate from \textcolor{red}{[L1]} to \textcolor{blue}{[L2]}: [TGT] \\

How do you say [SRC] in \textcolor{blue}{[L2]}? & translate from \textcolor{red}{[L1]} to \textcolor{blue}{[L2]}: [TGT] \\

[SRC] = \textcolor{blue}{[L2]}: &  translate from \textcolor{red}{[L1]} to \textcolor{blue}{[L2]}: [TGT] \\

Translate this from \textcolor{red}{[L1]} into \textcolor{blue}{[L2]}: [SRC] &  translate from \textcolor{red}{[L1]} to \textcolor{blue}{[L2]}: [TGT] \\

Translate this into \textcolor{blue}{[L2]}: [SRC] &  translate from \textcolor{red}{[L1]} to \textcolor{blue}{[L2]}: [TGT] \\

Given the following passage: [SRC], a good \textcolor{blue}{[L2]} translation is: & translate from \textcolor{red}{[L1]} to \textcolor{blue}{[L2]}: [TGT] \\

\textcolor{red}{[L1]}: [SRC] translates into \textcolor{blue}{[L2]} as: & translate from \textcolor{red}{[L1]} to \textcolor{blue}{[L2]}: [TGT] \\

If the \textcolor{red}{[L1]} version says: [SRC]; then the \textcolor{blue}{[L2]} version should say: &  translate from \textcolor{red}{[L1]} to \textcolor{blue}{[L2]}: [TGT] \\

What is the \textcolor{blue}{[L2]} translation of: [SRC]? & translate from \textcolor{red}{[L1]} to \textcolor{blue}{[L2]}: [TGT] \\
 
What is the \textcolor{blue}{[L2]} translation of the \textcolor{red}{[L1]} sentence: [SRC]? & translate from \textcolor{red}{[L1]} to \textcolor{blue}{[L2]}: [TGT] \\

\bottomrule
\end{tabular}
\end{center}
\end{table*}
\subsection{Task-enhanced Constrained Turning}
\label{para_tect}
The task-enhanced constrained turning pretends a translation task descriptive prefix to an encoder-decoder style or decoder-only transformer architecture to obtain LLM output $z$:
\begin{equation}
\label{equation_tect}
z=[P_{des}, y],
\end{equation}
where $[\cdot,\cdot]$ denotes the contact operation. Here, $P_{des}$ denotes a sequence of task descriptive prefix (\eg, $P_{des}$=translate from French to German:). 
Since each token in the prefix is in the vocabulary, we can directly initialize the $P_{des}$ based on the pre-trained weight matrix of LLM's vocabulary.
Table~\ref{table_instruction}, we provide a detailed list of the input/output format for instruction fine-turning.

\subsection{Auto-constriction Turning}
Different from task-enhanced constrained turning (\hardmodel), auto-constriction turning first generates the trigger tokes related to translation direction, and then generates the ground truth $y$:
\begin{equation}
\mathcal{L}=\sum_{(x,y) \sim D} \sum_{i \leq |z|} -\log p(z_{i} | x,z_{<i}; \theta),
\end{equation}
where $z=[t(x), y]$, and $t(\cdot)$ represents a mapping strategy that establishes an association between trigger token sequences and different translation directions.
Specifically, for translation task with different translation directions, we design two types of trigger tokens: \textit{common trigger token} and \textit{specific trigger token}. The common trigger token is designed to learn shared information in different translation directions, and the specific trigger token is to capture the target language information.
Therefore, each translation direction has common trigger tokens, and only the translation directions with the same target language have the same specific trigger token.
Both types of trigger tokens help the model understand instructions.

Trigger tokens are special tokens added to the vocabulary during fine-turning, which can also be seen as additional supervised information to determine whether the model understands the instructions.
Due to trigger tokens are added as special tokens, we can easily remove these trigger tokens during decoding.
% 它也可以被看作是额外的监督信息，判断模型是否理解了指令
We should note that the trigger tokens are generated on the target side. Therefore, the added trigger tokens can be combined and arranged arbitrarily, allowing them to represent even unseen semantic information during the fine-tuning stage.
As a result, auto-constriction turning performs well under zero-shot setting.

\section{Experiments}
% bleu的缺点: instead of BLEU which has been demonstrated to be suboptimal for high-quality translations[Prompting PaLM for Translation: Assessing Strategies and Performance]

\begin{table}[]
%\vspace{-4mm}
\caption{The basic statistics of training\&validation datasets.} 
\label{table_train_dataset}
%\vspace{-3mm}
\begin{center}
\setlength{\tabcolsep}{1.5mm}{} 
\begin{threeparttable}
\begin{tabular}{c|ccc}
\hline
Lang-Pair    & Training   & Validation   \\

\hline
EN-CS      & 2,000$\times$2          & 2,000$\times$2       \\
EN-DE      & 2,000$\times$2          & 2,000$\times$2       \\
EN-FR      & 2,000$\times$2          & 2,000$\times$2       \\
EN-ZH      & 2,000$\times$2          & 2,000$\times$2        \\
EN-RU      & 2,000$\times$2          & 2,000$\times$2       \\
EN-RO      & 2,000$\times$2          & 2,000$\times$2       \\
EN-UK      & 2,000$\times$2          & 2,000$\times$2        \\
EN-HI      & 2,000$\times$2          & 2,000$\times$2        \\
\hline
Total      & 32,000               & 32,000         \\
\hline
\end{tabular}
% \begin{tablenotes}
% \scriptsize
%  \item 
% \end{tablenotes}
\end{threeparttable} 
\end{center}
%\vspace{-8mm}
\end{table}
\begin{table}[]
%\vspace{-4mm}
\caption{The basic statistics of test sets. The arrow is the translation direction.} 
\label{table_test_dataset}
%\vspace{-3mm}
\begin{center}
\setlength{\tabcolsep}{1.5mm}{} 
\begin{threeparttable}
\begin{tabular}{c|c|c}
\hline
Benchmarks  & $\rightarrow$ &  $\leftarrow$  \\
\hline
WMT-22 EN-CS      &   2,037  & 1,448         \\
\hline
WMT-22 EN-DE      &2,037         & 1,984         \\
\hline
WMT-22 EN-ZH     &2,037         & 1,875         \\
\hline
WMT-22 EN-UK     &2,037          & 2,018         \\
\hline
WMT-22 EN-RU     &2,037        & 2,016         \\
\hline
WMT-22 DE-FR      &1,984          & 2,006         \\
\hline
WMT-22 CS-UK      &1,930          &  2,812        \\
\hline

WMT-16 EN-RO      &1,999         & 1,999         \\
\hline
WMT-14 EN-HI     &2,507         & 2,507         \\
\hline
WMT-14 EN-FR      &3,003         & 3,003         \\
\hline
\end{tabular}
\end{threeparttable} 
\end{center}
%\vspace{-8mm}
\end{table}

\subsection{Dataset}
Our training data always has either the source or the target in English~\cite{schioppa2023cross}. % We use the standard approach used when training multi-lingual supervised models % 为什么选择2k条数据
We select 8 language pairs in OPUS-100\footnote{https://object.pouta.csc.fi/OPUS-100/v1.0/opus-100-corpus-v1.0.tar.gz} for training and evaluating. To guarantee the quality of the training data, we use the LABSE \cite{feng2020language} tool to select sentences with similarity scores above 0.75.
Since our goal is to explore the translation ability of LLM and reduce the impact of parallel corpus on LLM, we limit the number of parallel sentences for training and validation to 2,000 per language pair, which follows the same settings as~\citet{li2023eliciting}. 
In order to minimize the possibility of overlap with mT0's training corpus and enable comparisons with state-of-the-art (SOTA) systems, we use the WMT 14, 16 and 22 benchmarks for evaluation. We evaluate our model and baselines on news data from the WMT 2022~\cite{vilar2022prompting, kocmi2022findings} evaluation campaign. WMT 22 moves away from testing only on the news domain like in previous years and shifts to focus on the general scenario covering news, social, conversational, and e-commerce.
Since Hindi, French and Romanian are common evaluated and not included in WMT 22, we use test data from WMT 14 and WMT 16.
Table~\ref{table_train_dataset} and Table~\ref{table_test_dataset} show statistics for our training, validation, and test data (since each language pair contains two directions, we multiply each cell by 2).

In the experiment, we find that although the selected training dataset for zh-en language pair has a high LABSE similarity score.
It often appears in both English and Chinese within the same sentence, which seriously damages the quality of training data. 
Hence, We translate this low-quality en-zh language pair, from English to Chinese, using the official ChatGPT API\footnote{https://platform.openai.com/docs/api-reference/chat} provided by OpenAI \cite{ouyang2022training}. 

To generalize instruction tuning, we collect publicly available multilingual translation prompts from~\citet{bawden2023investigating} and  PromptSource~\citet{bach2022promptsource} in Table~\ref{table_instruction}, and randomly select prompt and parallel sentence to combine as model input. Accordingly, the instruction tuning dataset is scrambled to ensure diversity and randomness. When doing the inference, we also randomly select the aforementioned translation prompts, in Table~\ref{table_instruction}, to generalize the evaluation results.

\subsection{Evaluation}
% To comprehensively evaluate the performance of multilingual machine translation, we utilize four widely-used metrics\footnote{If a source sentence has multiple references, we report the average score}.
To comprehensively evaluate the performance of multilingual machine translation, we utilize four widely-used metrics\footnote{If a source sentence has multiple references, we report the average score}: scaleBLEU \cite{post-2018-call}, chrF \cite{popovic2015chrf}, comet-da\footnote{https://github.com/Unbabel/COMET}, and perplexity (we restrict perplexity to the X-EN direction since GPT-2-Large \cite{radford2019language} has only been trained on English text).
Furthermore, to qualitatively determine the four types of translation errors mentioned in section \ref{introduction}, we apply the following metrics.
\begin{itemize}
    \item \textbf{OT ratio:} we utilize the \texttt{langdetect} package to detect the highest probability language of the generated translations. We calculate the proportion of the number of sentences that are translated into the wrong language out of all sentences.
    
    \item \textbf{SC ratio:} We detect source copy (SC) error by computing scareBLEU score between predictions and the source sentences. If scareBLEU score exceeds 80, we consider it to be a SC error.
    \item \textbf{OUG ratio:} We consider translations with a length ratio, compared to reference, exceeding 2 or falling below 0.5 as over/under generation (OUG) issue. Meanwhile, misunderstanding of instructions (MI) issue can also be reflected by sentence length. 
    % Therefore, we use this indicator to reflect the two types of errors A and B

\end{itemize}

\subsection{Implementation Details}

We train the baselines and \model on 8 Tesla-V100-32G GPUs.
For all the baselines, the random seed, learning rate, learning schedule, dropout rate, and gradient clipping, number beams, no repeat ngram size, number of common trigger tokens per translation direction, and number of specific trigger tokens per translation direction are set as 42, 2e-5, cosine, 0.1, 1.0, 5, 3, 1 and 1, respectively.
Due to memory limitations, the batch size is set to 2, the gradient accumulation step is set to 2, and training is performed with FP16.
The ratio of validation set used for evaluating is set to 0.1 empirically.
We adopt early stopping strategy, and the patience is set to 5.

mT0~\cite{muennighoff2022crosslingual} releases 5 variants of different sizes, ranging from 300M to 13B, referred to as "mT0-small" and "mT0-xxl" respectively. In this paper (where our goal is to design the recipe through extensive experimentation), we use mT0-xl to reduce computational costs. For all models and experiments, we use Hugging Face Transformers\footnote{https://github.com/huggingface/transformers}.
% Unless otherwise specified, we assume a default trigger length of 3 characters for each translation task.
\begin{table*}[]
%\vspace{-4mm}
\caption{Main results for mT0-xl on test datasets. Averages over different sets of translation directions (There are a total of 20 translation directions, including 8 from other languages to English, 8 from English to other languages, and 4 directions that do not involve English at all. Please refer to Table \ref{table_test_dataset} for details). } 
\label{table_main_result}
%\vspace{-3mm}
\begin{center}
\setlength{\tabcolsep}{1.5mm}{} 
\begin{threeparttable}
\begin{tabular}{c|cccccccc}
\hline
% & {Token-level}  & {Character-level} & {Semantic-level} & {Fluency-level} & \multicolumn{3}{c}{Error type (in $\%$)} \\
&   &  & &  &  & \multicolumn{3}{c}{Error type (in $\%$)} \\
Baselines    & Size &  scareBLEU &  chrF  & comet-da & perplexity  & OT ratio  &  SC ratio & OUG ratio\\

\hline
mT0-xl 0-shot &   3.7B  & 16.06  & 35.59  & 72.87 & 175.57 & 32.80 & 12.52 & 6.06\\
mT0-xl 1-shot  &  3.7B  & 7.29   & 21.28  & 55.65 & \textbf{83.60} & 39.58 & 23.60 & 22.18 \\
mT0-xxl 0-shot &  13B   & 21.06 & 43.08 & 77.70 & 270.00 & 17.77 & 4.82 & 2.77\\
mFTI     & 3.7B  & 25.50  & 50.17  & 82.13 & 144.49 & 6.26 & 0.70 & 0.88  \\
% MPT      &            &          \\

\hline
\hardmodel & 3.7B &\textbf{25.94}  & 50.61  & \textbf{82.40} & 242.71 & 4.19 & 0.29 & 0.75 \\
\model    & 3.7B &25.87 & \textbf{50.65} & 82.39 & 126.25 & \textbf{3.91} & \textbf{0.19} & \textbf{0.58}     \\
\hline
\end{tabular}
% \begin{tablenotes}
% \scriptsize
%  \item 
% \end{tablenotes}
\end{threeparttable} 
\end{center}
%\vspace{-8mm}
\end{table*}
\subsection{Baselines}
We compare our method with following baselines: 
\begin{itemize}
    \item \textbf{0-shot:} In-context learning (ICL) \cite{brown2020language} is a powerful method for harnessing the capabilities of LLM, which provides a few demonstrations to guide models on how to perform even on an unseen task without fine-tuning. 

    \item \textbf{1-shot:} To guarantee the quality of demonstrations, we randomly select demonstrations from the FLORES-200 \cite{costa2022no} validation dataset for the 1-shot setting.
    We do not apply more demonstrations, since more demonstrations do not lead to  higher performance in case of mT0 \cite{muennighoff2022crosslingual}. 
    % \item \textbf{PromptTuning:} \cite{lester2021power} is a competitive parameter-efficient fine-tuning technique for adapting frozen pretrained language models to downstream tasks.
    
    \item \textbf{mFTI:} \cite{li2023eliciting} is a full fine-tuning method, where all model parameters are tuned during adaptation on multilingual machine translation task with a limited training samples.
    It can also be regarded as instruction fine-tuning.

    % \item \textbf{MPT:} multitask prompt tuning (MPT) \cite{wang2023multitask} distills knowledge from multiple task-specific source prompts firstly, and then apply multiplicative low-rank updates to efficiently adapt task-shared prompt to each downstream target task.

    \item \textbf{\hardmodel:} is a variant of our proposed model mentioned in Equation \ref{equation_tect}. 
    It utilizes a hard encoding constrained template in the target side to guide the generation of the LLM (hard encoding constrained template is empirically set to {\textit{translate from [L1] to [L2]}}.
    
    % which utilises a hard encoding constrained template (hard encoding constrained template is empirically set to \texttt{translate from [L1] to [L2]:\footnote{where [L1] denotes source language, [L2] denotes the target language.}} in experiments) in the target side to guide the generation of LLM.

\end{itemize}

\subsection{Experimental Results}

The average results on both zero-shot and supervised directions, along with other baselines, are shown in Table \ref{table_main_result}, where the best result is displayed in bold.
% The average results on both zero-shot and supervised translation directions with other baselines are shown in Table \ref{table_main_result}, where the best is shown in bold.
According to Table \ref{table_main_result}, we can find that both \hardmodel and \model outperform all baselines in terms of all evaluation metrics at all levels (except the perplexity), based on the average results for all translation directions, and even outperforms larger models, such as mT0-xxl 13B.
Specifically, \hardmodel improves 0.44, and \model improves 0.37 in scareBLEU compared to the bested performed baseline, mFTI.
Furthermore, \model achieves the minimum OT ratio, SC ratio and OUG ratio, as well as the maximum chrF-style score. 
These results indicate that the auto-constriction template can assist LLM in better understanding instructions and minimize translation issues (\eg, misunderstanding of instructions issue, off-target translation issue,  source copy issue and over/under generation issue).
The main reason is that our designed constrained template on the target side can extract trigger tokens as additional supervised information.

In addition, we can observe that mT0-xl exhibits a weak capacity for in-context learning. 1-shot baseline achieves lower performance compared to 0-shot baseline, which is consistent with~\citet{muennighoff2022crosslingual}.
We also find that \model shows slightly weaker but comparable performance, \eg on semantic-level metrics, compared to \hardmodel.
We consider that hard encoding provides a clearer guidance. However, designing an optimal constraint template requires profound insights into the task information.
% 建立语言方向和翻译结果得一种关系
% Although the constraint from \model is not grammatical ("{sub} ediatric striker ice baseman defensive {obj}"), it does contain tokens that are directly related to the prompt.
\begin{table}[]
%\vspace{-4mm}
\caption{Five constrained templates used in the ablation experiment.} 
\label{table_factor_templete}
%\vspace{-3mm}
\begin{center}
\setlength{\tabcolsep}{1.5mm}{} 
\begin{threeparttable}
\begin{tabular}{c|c}
\hline

Variants    & Constrained Template      \\

\hline
% \hardmodel-0      & Task is translate from [L1] to [L2]:        \\
\hardmodel-1      & translate from [L1] to [L2]:          \\
\hardmodel-2      & translate to [L2]:           \\
\hardmodel-3      & translate from [L1]:           \\
\hardmodel-4      & from [L1] to [L2]:            \\
\hardmodel-5      & [L2]:             \\
\hline
\end{tabular}
% \begin{tablenotes}
% \scriptsize
%  \item 
% \end{tablenotes}
\end{threeparttable} 
\end{center}
%\vspace{-8mm}
\end{table}
\begin{table*}[]
%\vspace{-4mm}
\caption{Averages over different sets of translation directions with five different constrained templates.} 
\label{table_factor_result}
%\vspace{-3mm}
\begin{center}
\setlength{\tabcolsep}{1.5mm}{} 
\begin{threeparttable}
\begin{tabular}{c|ccccccc}
\hline
% & \multicolumn{2}{c}{Token-level}  & \multicolumn{3}{c}{Character-level} & \multicolumn{1}{c}{Semantic-level} & \multicolumn{1}{c}{Fluency-level} & \multicolumn{3}{c}{Error type (in $\%$)} \\
&   &  &  &  & \multicolumn{3}{c}{Error type (in $\%$)} \\
baselines    &  scareBLEU   &  chrF & comet-da & perplexity  & OT ratio &  SC ratio & OUG ratio\\

\hline
% \hardmodel-0      & 25.93  & 50.54  & 82.39 & 175.57 & 4.30 & 0.32 & 0.79 \\
\hardmodel-1      & 25.94  & 50.61  & 82.40 & 242.71 & 4.19 & 0.29 & 0.75 \\
\hardmodel-2      & \textbf{26.11}  & \textbf{50.83}  & 82.49 & 134.80 & \textbf{4.17} & 0.32 & 0.71 \\
\hardmodel-3      & 26.02  & 50.76  & 82.45 & 128.62 & 4.18 & 0.33 & 0.72 \\
\hardmodel-4      & \textbf{26.11}  & \textbf{50.83}  & \textbf{82.57} & \textbf{127.62} & 4.27 & \textbf{0.21} & \textbf{0.66} \\
\hardmodel-5      & 25.98  & 50.65 & 82.45 & 129.67 & 4.42 & 0.31 & 0.72\\
\hline

\end{tabular}
% \begin{tablenotes}
% \scriptsize
%  \item 
% \end{tablenotes}
\end{threeparttable} 
\end{center}
%\vspace{-8mm}
\end{table*}

% \begin{table*}[]
% \caption{Averages over different sets of translation directions. Parameter sensitivity \wrt ~number of trigger tokens.} 
% \label{table_trigger_result}
% \begin{center}
% \setlength{\tabcolsep}{1.5mm}{} 
% \begin{threeparttable}
% \begin{tabular}{c|ccccccc}
% \hline
% &   &  &  &  & \multicolumn{3}{c}{Error type (in $\%$)} \\
% $\#$ trigger tokens    &  scareBLEU   &  chrF  & comet-da & perplexity  & OT ratio &  SC ratio & OUG ratio\\
% \hline
% 1      & \textbf{25.75} & 50.52 & \textbf{82.40} & 145.65 & 3.97 & 0.22 & 0.82\\
% 3      & 25.74 & \textbf{50.56} & 82.35 & 122.88 & \textbf{3.59} & 0.19 & \textbf{0.61} \\
% 6      & 25.60 & 50.38 & 82.18 & 120.47 & 3.69 & 0.23 & 0.69\\
% 9      & 25.69 & 50.51 & 82.15 & \textbf{111.88} & 3.75 & \textbf{0.16} & \textbf{0.61} \\
% \hline
% \end{tabular}
% \end{threeparttable} 
% \end{center}
% \end{table*}

\begin{table*}[]
\caption{Parameter sensitivity \wrt ~number of trigger tokens.} 
\label{table_trigger_result}
\begin{center}
\setlength{\tabcolsep}{1.5mm}{} 
\begin{threeparttable}
\begin{tabular}{cc|ccccccc}
\hline
\multicolumn{2}{c}{$\#$ trigger tokens} &   &  &  &  & \multicolumn{3}{c}{Error type (in $\%$)} \\
common & specific  &  scareBLEU   &  chrF  & comet-da & perplexity  & OT ratio &  SC ratio & OUG ratio\\
\hline
1 & 1   & \textbf{25.87} & \textbf{50.65} & \textbf{82.39} & 126.25 & 3.91 & \textbf{0.19} & \textbf{0.58}     \\
1 & 3   & 25.80 & 50.44 & 82.23 & \textbf{119.26} & 3.96 & 0.40 & 0.69\\
1 & 6   & 25.78 & 50.52 & 82.25 & 128.73 & \textbf{3.67} & 0.23 & 0.71\\
1 & 9   & 25.31 & 49.56 & 81.73 & 124.52 & 4.94 & 0.24 & 1.96\\
2 & 6   & 25.55 & 50.25 & 82.05 & 121.29 & 3.74 & 0.24 & 0.79\\
3 & 3   & 25.32  & 49.96  & 81.79  & 121.80 & 3.98 & 0.24 & 0.85\\
3 & 9   & 24.09 & 46.90 & 79.81 & 121.78 & 8.75 & 0.34 & 5.60 \\
6 & 6   &  24.18 & 47.23  & 79.88  & 134.70 & 8.37 & 0.49 & 5.06  \\
9 & 9   &  23.95 & 46.44  & 78.83  & 122.77 & 9.83 & 0.48 & 6.84\\
\hline
\end{tabular}
\end{threeparttable} 
\end{center}
\end{table*}
\subsection{Ablation Study}
\subsubsection{Ablation Experiment on Constrained Template Component}
To determine the impact of each part of the constrained template on the translation performance, we design five variations of the constrained template elaborately, as shown in Table~\ref{table_factor_templete}.
The results of ablation study are presented in Table~\ref{table_factor_result}. 

First, as we can see that all \hardmodel based methods outperform the instruction fine-tuning baseline, mFTI, which indicates that the constrained template can help LLM understand instructions, alleviate the four translation errors mentioned above, and guide the generation of the optimal outputs.
% Second, \hardmodel-1 achieves better performance than \hardmodel-0, which indicates some common string, \eg \texttt{Task is} is redundant information and will lead to a decrease in the performance.
Second, \hardmodel-2 achieves better performance than \hardmodel-3.
It shows that the target language information plays a more important role than the source language information, which in line with~\citet{wu2021language}.
Third, \hardmodel-4 achieves better performance than \hardmodel-5, which indicates that including only target language information is not sufficient, and understanding how to perform translation tasks is equally important.
Overall, we have the conclusion that providing information about task execution helps guide the model's output and improves performance.
\begin{figure}[t]
\centering
\includegraphics[width=0.4\textwidth]{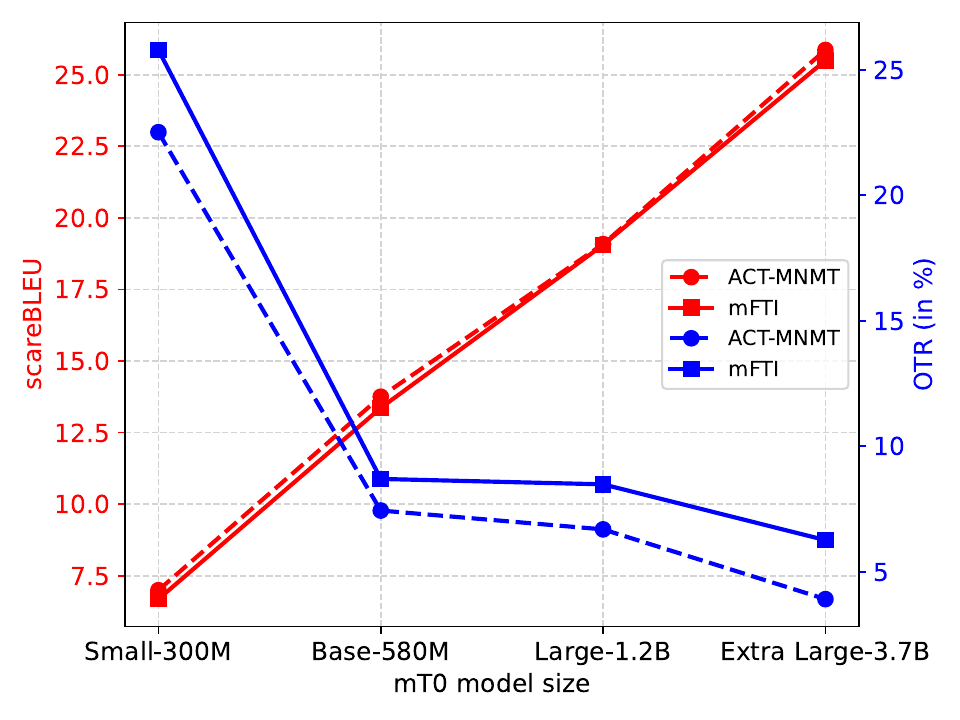} % Reduce the figure size so that it is slightly narrower than the column.
\caption{Parameter sensitivity \wrt ~ model size.}
\label{fig_model_size_blue_otr}
\end{figure}

% the key components all contribute to performance improvement of our model.

\subsubsection{Effect of Number of Triggers}
We show the  result in Table \ref{table_trigger_result}.
We observe that as the number of common trigger tokens and specific trigger tokens increases, the performance of the model gradually decreases.
% On the on hand, when we fix the number of common trigger tokens and increase the specific trigger tokens
In addition, when analyzing the experimental results in depth, we observe that perplexity is highly sensitive in the translation direction from Chinese to English.
Specifically, when the number of common trigger token and specific trigger token change from 1 to 3, the perplexity in the Chinese to English direction changes from 234.89 changes 180.75.
This greatly reduces the average improvement in other translation directions.

%We use red lines to represent scaleBLEU and blue lines to represent the off-target ratio.

\begin{figure*}[h]
	\centering 
% 	\vspace{-6mm}
% 	\hspace{-5mm} 
	\subfigure[average]{
	    \label{fig_data_size_avg}
		\includegraphics[width=.33\linewidth]{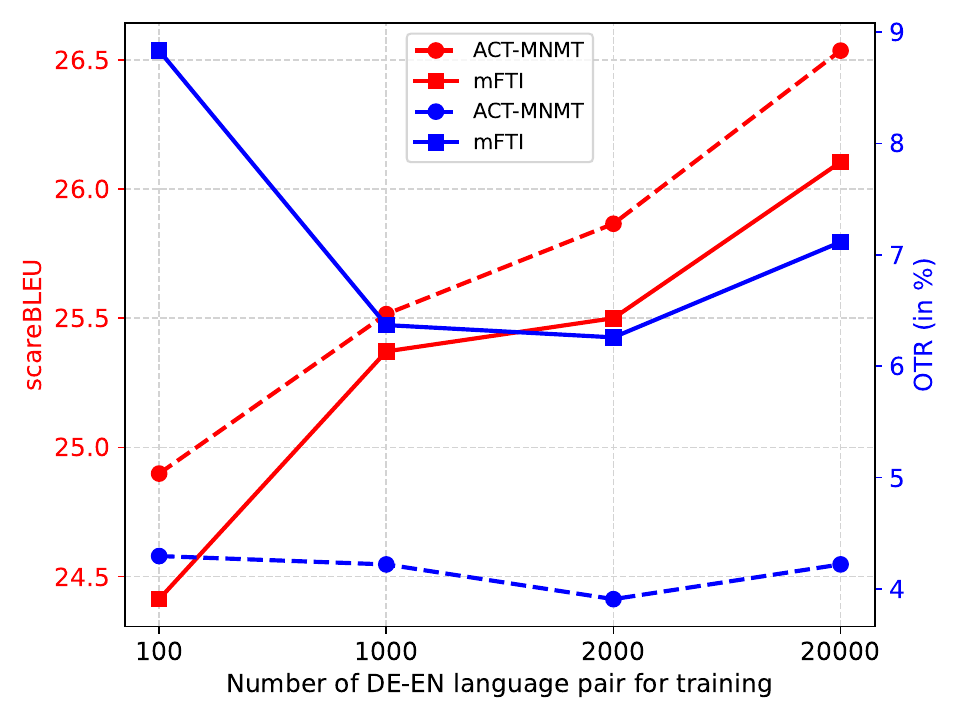} 
	} 
% 	\quad{}  
	\hspace{-5mm}
	\subfigure[DE to EN]{
	    \label{fig_data_size_deen}
		\includegraphics[width=.33\linewidth]{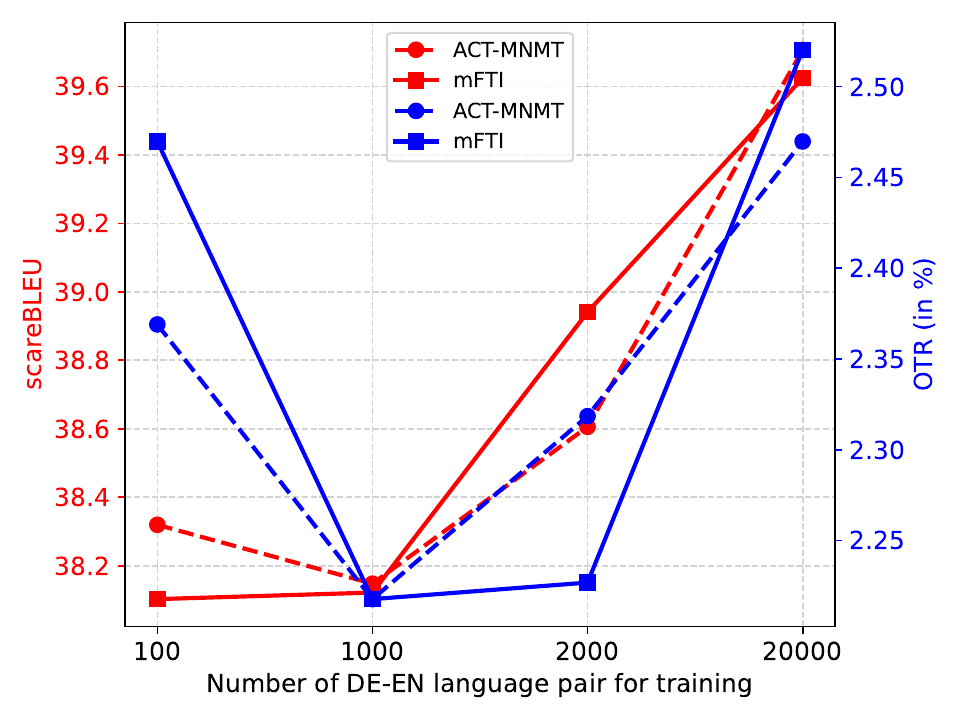}
	}
	\hspace{-5mm}
	\subfigure[EN to DE]{
	    \label{fig_data_size_ende}
		\includegraphics[width=.33\linewidth]{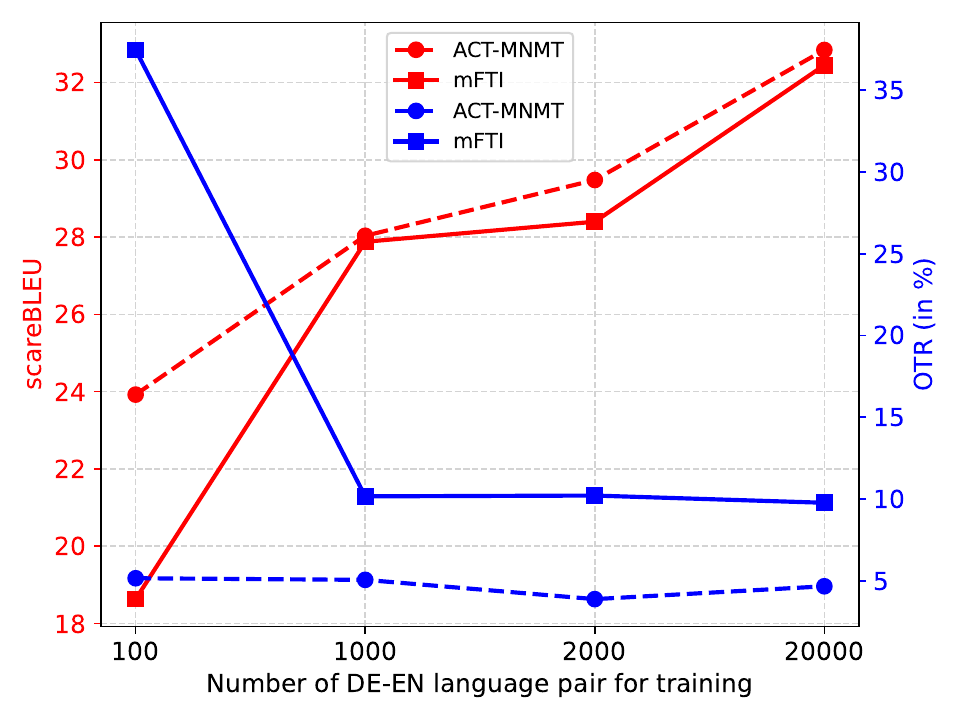}
	} 
% 	\vspace{-2mm}
	\caption{Parameter sensitivity \wrt ~ number of DE-EN language pair for training.}
	\label{fig_data_size}
% 	\vspace{-3mm}
\end{figure*}

\subsubsection{Effect of Model Size} % % Data Scalbility

Since there exist different versions of mT0 with differing numbers of parameters, we perform the experiment to determine whether the \model has a good scalability and how model size impacts the performance. In this experiment, we report the average scaleBLEU score and off-target ratio across 20 translation directions in Figure~\ref{fig_model_size_blue_otr}.

Some observations are summarized from the experimental results:
(1) Our approach, \model, can be applied to different model sizes and achieve higher scaleBLEU and lower off-target ratio compared to traditional instruction fine-tuning. % 
(2) \model can effectively alleviate off-target issue. Specifically, \model obtains 12.71$\%$, 14.48$\%$, 21.58$\%$ and 37.55$\%$ reduction compared to mFTI in terms of Small-300M, Base-580M, Large-1.2B and Extra Large-3.7B on off-target ratio metric, respectively, and achieves equal or even better scareBLE score. 
Furthermore, as the model parameters increase, \model consistently outperforms mFTI under the same parameters, which shows the robustness of \model.

\subsubsection{Effect of Data Size} % Data Scalbility

In this subsection, we will investigate how the number of training examples affects the model.
We only change the number of training language pairs for DE-EN (German-English), and keep the number of other language pairs unchanged.
Figure \ref{fig_data_size} presents scareBLEU and off-target ratio (OTR) in terms of average over all directions, DE to EN direction and EN to DE direction, respectively.

Firstly, we observe a common phenomenon in the three subgraphs of Figure \ref{fig_data_size}, where the OTR shows a trend of initially decreasing and then increasing with the increase in the number of DE-EN language pairs (the knee point may be at 1,000 or 2,000).
This phenomenon can be explained by the fact that when the number of German examples exceeds 2,000, it can cause an imbalance in multilingual language pairs, leading to an increased probability of translating into German.
Secondly, \model exhibits strong data scalability, which significantly improves the translation quality, especially when there are limited DE-EN language pairs for training.
For instance, in the case of 100 English German language pairs, \model obtains 5.29 scaleBLEU improvement, and achieve 86.24$\%$ decrease on OTR metric compared to mFTI.
In addition, as English and German increased from 100 to 20,000, DE to EN increases by 1.39 scareBLEU and EN to DE increased by 8.92 scareBLEU, which indicates that translating  from other languages to German is more sensitive than translate from German to other languages.

\subsubsection{Over/Under-generation}
\label{auxiliary_experiments}
\begin{figure}[h]
	\centering 
% 	\vspace{-6mm}
% 	\hspace{-5mm} 
	\subfigure[BLOOMZ-7B1]{
	    \label{fig_bloomz7b1_oug}
		\includegraphics[width=.51\linewidth]{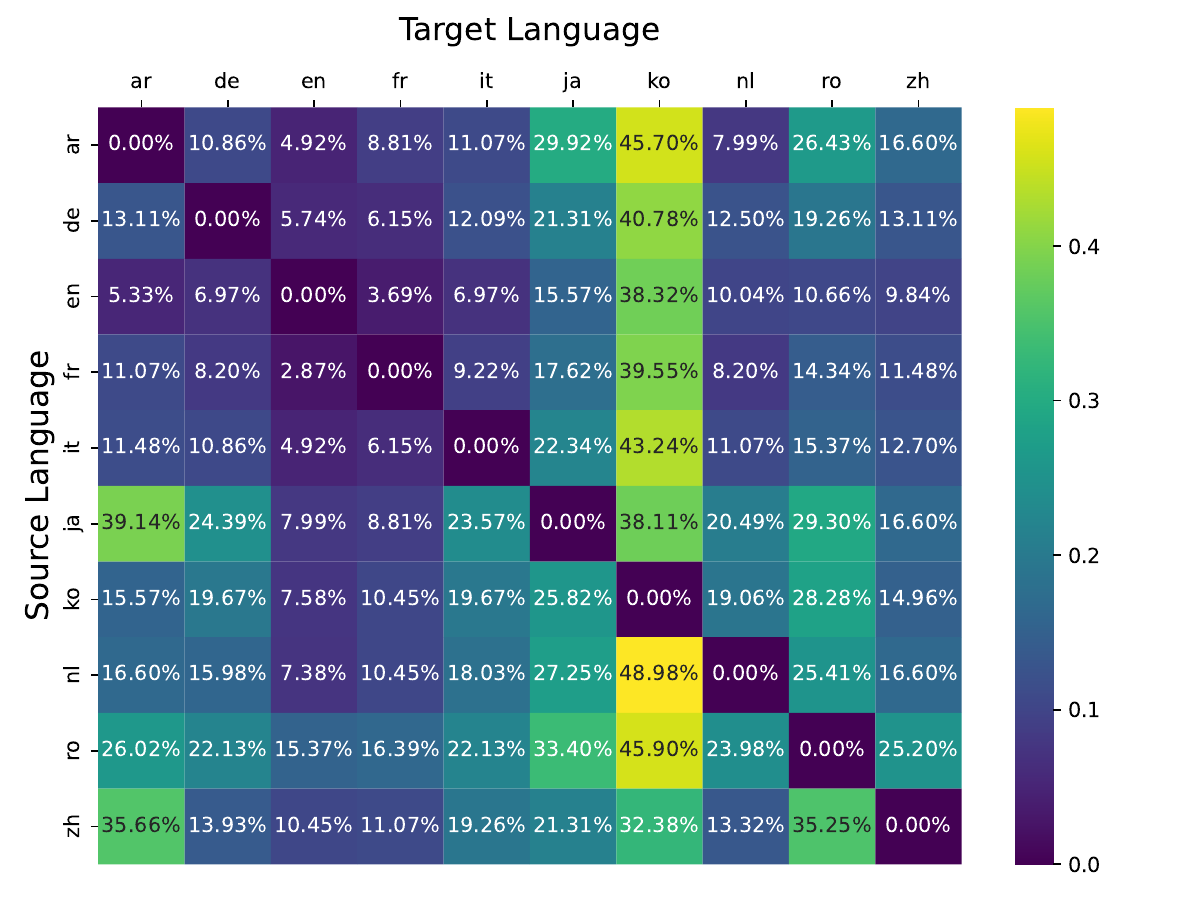} 
	} 
% 	\quad{}  
	\hspace{-9mm}
	\subfigure[mT0-xl (3.7B)]{
	    \label{fig_mt0xl_oug}
		\includegraphics[width=.51\linewidth]{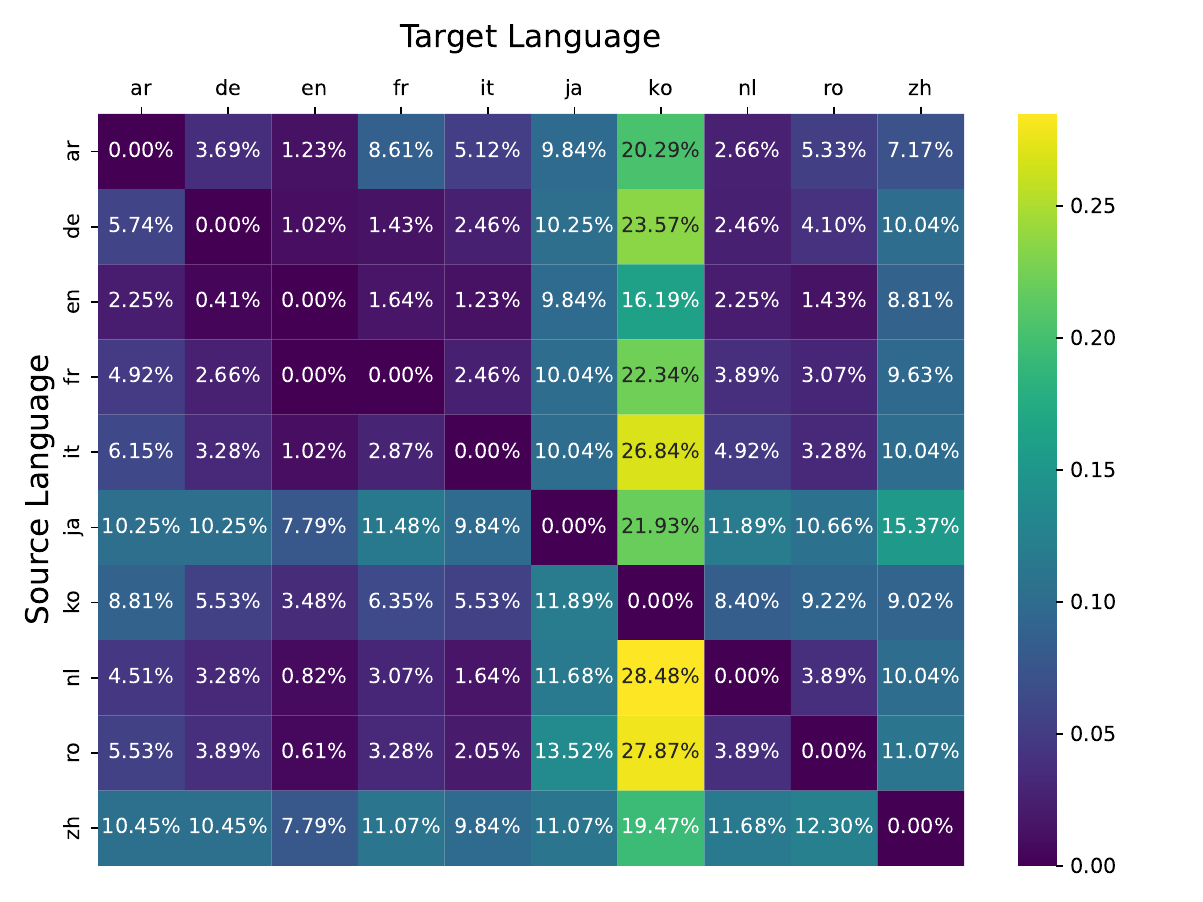}
	} 
% 	\vspace{-2mm}
	\caption{Over/Under-generation ratio on IWSLT 2017 test datasets.}
	\label{fig_oug}
% 	\vspace{-3mm}
\end{figure}
Figure~\ref{fig_oug} reports the over/under-generation ratio of any pair of languages on IWSLT 2017 test datasets.
% We classify the translations with a length ratio exceeding 2 or falling below 0.5 as over- or under-translations, correspondingly. 
Since BLOOMZ and mT0 have trained on FLORES-200~\cite{muennighoff2022crosslingual}, we do not evaluate on this standard multilingual translation dataset. As illustrated in Figure \ref{fig_oug}, we have the following observations: 1) Over/Under-generation (OUG) ratio and off-target (OT) ratio have the similar distributions among language pairs. For instance, both the OUG and OT ratio exhibit small values among rich-resource language pairs, and present large values among low-resource language pairs. 2) Compared to OT ratio, OUG ratio is lower among all directions, which means OT issue is more severe than OUG issue.
% 3) mT0-xl shows a better translation quality compared to BLOOMZ-7B1, even though the later has more parameters.
% \subsection{Zero-Shot Performance}

% \model scale up We now investigate the sensitivity of our model 

% \subsubsection{Effect of Corpus Quality} % Data Scalbility

\section{Conclusion}
In this paper, we first propose a novel task-enhanced constrained turning (\hardmodel) method, which utilises a manually designed hard encoding constrained template to guide the output of the model.
% However, writing specific descriptive information is not only time consuming, but it is not clear that the same phrasing will be effective for every model. 
Then, we extend \hardmodel to an auto-constriction turning method (\model), which applies a sequence of trigger token as a soft-encoding constrained template to guide the output of the model.
Auto-constriction template can effectively learn patterns in different translation directions, alleviate OT problems, OUG problems, and SC problems.
Both \hardmodel and \model significantly outperforms instruction fine-tuning baseline.
We demonstrate that the \model model has good scalability and robustness through extensive experiments, and conclude that providing information about task execution helps guide the model’s generation and improves performance.
Although we focus only on mT0 model in this work, our method is trivially extendable to auto-regressive language models. 

The main limitation of our work is that although we reduce the cost of training by reducing the number of training samples, we still need to fine-tune the weights of the original model, which can be troublesome in some scenarios.
In the future, we will be devoted to using more efficient methods to solve off-target issue in multilingual machine translation.

\nocite{*}
\section{Bibliographical References}\label{sec:reference}

\bibliographystyle{lrec-coling2024-natbib}
\bibliography{lrec-coling2024-example}

\begin{thebibliography}{65}
\expandafter\ifx\csname natexlab\endcsname\relax\def\natexlab#1{#1}\fi

\bibitem[{Agrawal et~al.(2022)Agrawal, Zhou, Lewis, Zettlemoyer, and
  Ghazvininejad}]{agrawal2022context}
Sweta Agrawal, Chunting Zhou, Mike Lewis, Luke Zettlemoyer, and Marjan
  Ghazvininejad. 2022.
\newblock In-context examples selection for machine translation.
\newblock \emph{arXiv preprint arXiv:2212.02437}.

\bibitem[{Bach et~al.(2022)Bach, Sanh, Yong, Webson, Raffel, Nayak, Sharma,
  Kim, Bari, Fevry, Alyafeai, Dey, Santilli, Sun, Ben-David, Xu, Chhablani,
  Wang, Fries, Al-shaibani, Sharma, Thakker, Almubarak, Tang, Tang, Jiang, and
  Rush}]{bach2022promptsource}
Stephen~H. Bach, Victor Sanh, Zheng-Xin Yong, Albert Webson, Colin Raffel,
  Nihal~V. Nayak, Abheesht Sharma, Taewoon Kim, M~Saiful Bari, Thibault Fevry,
  Zaid Alyafeai, Manan Dey, Andrea Santilli, Zhiqing Sun, Srulik Ben-David,
  Canwen Xu, Gunjan Chhablani, Han Wang, Jason~Alan Fries, Maged~S.
  Al-shaibani, Shanya Sharma, Urmish Thakker, Khalid Almubarak, Xiangru Tang,
  Xiangru Tang, Mike Tian-Jian Jiang, and Alexander~M. Rush. 2022.
\newblock \href {http://arxiv.org/abs/2202.01279} {Promptsource: An integrated
  development environment and repository for natural language prompts}.

\bibitem[{Bawden and Yvon(2023)}]{bawden2023investigating}
Rachel Bawden and Fran{\c{c}}ois Yvon. 2023.
\newblock Investigating the translation performance of a large multilingual
  language model: the case of bloom.
\newblock \emph{arXiv preprint arXiv:2303.01911}.

\bibitem[{Brown et~al.(2020)Brown, Mann, Ryder, Subbiah, Kaplan, Dhariwal,
  Neelakantan, Shyam, Sastry, Askell et~al.}]{brown2020language}
Tom Brown, Benjamin Mann, Nick Ryder, Melanie Subbiah, Jared~D Kaplan, Prafulla
  Dhariwal, Arvind Neelakantan, Pranav Shyam, Girish Sastry, Amanda Askell,
  et~al. 2020.
\newblock Language models are few-shot learners.
\newblock \emph{Advances in neural information processing systems},
  33:1877--1901.

\bibitem[{BSI(1973{\natexlab{a}})}]{bs-2570-manual}
BSI. 1973{\natexlab{a}}.
\newblock \emph{Natural Fibre Twines}, 3rd edition.
\newblock British Standards Institution, London.
\newblock BS 2570.

\bibitem[{BSI(1973{\natexlab{b}})}]{bs-2570-techreport}
BSI. 1973{\natexlab{b}}.
\newblock Natural fibre twines.
\newblock BS 2570, British Standards Institution, London.
\newblock 3rd. edn.

\bibitem[{Castor and Pollux(1992)}]{CastorPollux-92}
A.~Castor and L.~E. Pollux. 1992.
\newblock The use of user modelling to guide inference and learning.
\newblock \emph{Applied Intelligence}, 2(1):37--53.

\bibitem[{Cettolo et~al.(2017)Cettolo, Federico, Bentivogli, Jan, Sebastian,
  Katsuitho, Koichiro, and Christian}]{cettolo2017overview}
Mauro Cettolo, Marcello Federico, Luisa Bentivogli, Niehues Jan, St{\"u}ker
  Sebastian, Sudoh Katsuitho, Yoshino Koichiro, and Federmann Christian. 2017.
\newblock Overview of the iwslt 2017 evaluation campaign.
\newblock In \emph{Proceedings of the 14th International Workshop on Spoken
  Language Translation}, pages 2--14.

\bibitem[{Chen et~al.(2023{\natexlab{a}})Chen, Ma, Zhang, Wei, and
  Chang}]{Chen_Ma_Zhang_Wei_Chang_2023}
Liang Chen, Shuming Ma, Dongdong Zhang, Furu Wei, and Baobao Chang.
  2023{\natexlab{a}}.
\newblock On the off-target problem of zero-shot multilingual neural machine
  translation.

\bibitem[{Chen et~al.(2023{\natexlab{b}})Chen, Ma, Zhang, Wei, and
  Chang}]{chen2023off}
Liang Chen, Shuming Ma, Dongdong Zhang, Furu Wei, and Baobao Chang.
  2023{\natexlab{b}}.
\newblock On the off-target problem of zero-shot multilingual neural machine
  translation.
\newblock \emph{arXiv preprint arXiv:2305.10930}.

\bibitem[{Chercheur(1994)}]{Chercheur-94}
J.L. Chercheur. 1994.
\newblock \emph{Case-Based Reasoning}, 2nd edition.
\newblock Morgan Kaufman Publishers, San Mateo, CA.

\bibitem[{Chomsky(1973)}]{chomsky-73}
N.~Chomsky. 1973.
\newblock Conditions on transformations.
\newblock In \emph{A festschrift for {Morris Halle}}, New York. Holt, Rinehart
  \& Winston.

\bibitem[{Chowdhery et~al.(2022)Chowdhery, Narang, Devlin, Bosma, Mishra,
  Roberts, Barham, Chung, Sutton, Gehrmann et~al.}]{chowdhery2022palm}
Aakanksha Chowdhery, Sharan Narang, Jacob Devlin, Maarten Bosma, Gaurav Mishra,
  Adam Roberts, Paul Barham, Hyung~Won Chung, Charles Sutton, Sebastian
  Gehrmann, et~al. 2022.
\newblock Palm: Scaling language modeling with pathways.
\newblock \emph{arXiv preprint arXiv:2204.02311}.

\bibitem[{Chung et~al.(2022)Chung, Hou, Longpre, Zoph, Tay, Fedus, Li, Wang,
  Dehghani, Brahma et~al.}]{chung2022scaling}
Hyung~Won Chung, Le~Hou, Shayne Longpre, Barret Zoph, Yi~Tay, William Fedus,
  Eric Li, Xuezhi Wang, Mostafa Dehghani, Siddhartha Brahma, et~al. 2022.
\newblock Scaling instruction-finetuned language models.
\newblock \emph{arXiv preprint arXiv:2210.11416}.

\bibitem[{Costa-juss{\`a} et~al.(2022)Costa-juss{\`a}, Cross, {\c{C}}elebi,
  Elbayad, Heafield, Heffernan, Kalbassi, Lam, Licht, Maillard
  et~al.}]{costa2022no}
Marta~R Costa-juss{\`a}, James Cross, Onur {\c{C}}elebi, Maha Elbayad, Kenneth
  Heafield, Kevin Heffernan, Elahe Kalbassi, Janice Lam, Daniel Licht, Jean
  Maillard, et~al. 2022.
\newblock No language left behind: Scaling human-centered machine translation.
\newblock \emph{arXiv preprint arXiv:2207.04672}.

\bibitem[{Du et~al.(2021)Du, Qian, Liu, Ding, Qiu, Yang, and Tang}]{du2021glm}
Zhengxiao Du, Yujie Qian, Xiao Liu, Ming Ding, Jiezhong Qiu, Zhilin Yang, and
  Jie Tang. 2021.
\newblock Glm: General language model pretraining with autoregressive blank
  infilling.
\newblock \emph{arXiv preprint arXiv:2103.10360}.

\bibitem[{Ebrahimi et~al.(2017)Ebrahimi, Rao, Lowd, and
  Dou}]{ebrahimi2017hotflip}
Javid Ebrahimi, Anyi Rao, Daniel Lowd, and Dejing Dou. 2017.
\newblock Hotflip: White-box adversarial examples for text classification.
\newblock \emph{arXiv preprint arXiv:1712.06751}.

\bibitem[{Eco(1990)}]{Eco:1990}
Umberto Eco. 1990.
\newblock \emph{The Limits of Interpretation}.
\newblock Indian University Press.

\bibitem[{Fan et~al.(2021)Fan, Bhosale, Schwenk, Ma, El-Kishky, Goyal, Baines,
  Celebi, Wenzek, Chaudhary et~al.}]{fan2021beyond}
Angela Fan, Shruti Bhosale, Holger Schwenk, Zhiyi Ma, Ahmed El-Kishky,
  Siddharth Goyal, Mandeep Baines, Onur Celebi, Guillaume Wenzek, Vishrav
  Chaudhary, et~al. 2021.
\newblock Beyond english-centric multilingual machine translation.
\newblock \emph{The Journal of Machine Learning Research}, 22(1):4839--4886.

\bibitem[{Farhad et~al.(2021)Farhad, Arkady, Magdalena, Ond{\v{r}}ej, Rajen,
  Vishrav, Costa-jussa, Cristina, Angela, Christian
  et~al.}]{farhad2021findings}
Akhbardeh Farhad, Arkhangorodsky Arkady, Biesialska Magdalena, Bojar
  Ond{\v{r}}ej, Chatterjee Rajen, Chaudhary Vishrav, Marta~R Costa-jussa,
  Espa{\~n}a-Bonet Cristina, Fan Angela, Federmann Christian, et~al. 2021.
\newblock Findings of the 2021 conference on machine translation (wmt21).
\newblock In \emph{Proceedings of the Sixth Conference on Machine Translation},
  pages 1--88. Association for Computational Linguistics.

\bibitem[{Feng et~al.(2020)Feng, Yang, Cer, Arivazhagan, and
  Wang}]{feng2020language}
Fangxiaoyu Feng, Yinfei Yang, Daniel Cer, Naveen Arivazhagan, and Wei Wang.
  2020.
\newblock Language-agnostic bert sentence embedding.
\newblock \emph{arXiv preprint arXiv:2007.01852}.

\bibitem[{Freitag et~al.(2022)Freitag, Rei, Mathur, Lo, Stewart, Avramidis,
  Kocmi, Foster, Lavie, and Martins}]{freitag2022results}
Markus Freitag, Ricardo Rei, Nitika Mathur, Chi-kiu Lo, Craig Stewart,
  Eleftherios Avramidis, Tom Kocmi, George Foster, Alon Lavie, and Andr{\'e}~FT
  Martins. 2022.
\newblock Results of wmt22 metrics shared task: Stop using bleu--neural metrics
  are better and more robust.
\newblock In \emph{Proceedings of the Seventh Conference on Machine Translation
  (WMT)}, pages 46--68.

\bibitem[{Goyal et~al.(2022)Goyal, Gao, Chaudhary, Chen, Wenzek, Ju, Krishnan,
  Ranzato, Guzm{\'a}n, and Fan}]{goyal2022flores}
Naman Goyal, Cynthia Gao, Vishrav Chaudhary, Peng-Jen Chen, Guillaume Wenzek,
  Da~Ju, Sanjana Krishnan, Marc’Aurelio Ranzato, Francisco Guzm{\'a}n, and
  Angela Fan. 2022.
\newblock The flores-101 evaluation benchmark for low-resource and multilingual
  machine translation.
\newblock \emph{Transactions of the Association for Computational Linguistics},
  10:522--538.

\bibitem[{Hendy et~al.(2023)Hendy, Abdelrehim, Sharaf, Raunak, Gabr,
  Matsushita, Kim, Afify, and Awadalla}]{hendy2023good}
Amr Hendy, Mohamed Abdelrehim, Amr Sharaf, Vikas Raunak, Mohamed Gabr, Hitokazu
  Matsushita, Young~Jin Kim, Mohamed Afify, and Hany~Hassan Awadalla. 2023.
\newblock How good are gpt models at machine translation? a comprehensive
  evaluation.
\newblock \emph{arXiv preprint arXiv:2302.09210}.

\bibitem[{Hoel(1971{\natexlab{a}})}]{hoel-71-whole}
Paul~Gerhard Hoel. 1971{\natexlab{a}}.
\newblock \emph{Elementary Statistics}, 3rd edition.
\newblock Wiley series in probability and mathematical statistics. Wiley, New
  York, Chichester.
\newblock ISBN 0~471~40300.

\bibitem[{Hoel(1971{\natexlab{b}})}]{hoel-71-portion}
Paul~Gerhard Hoel. 1971{\natexlab{b}}.
\newblock \emph{Elementary Statistics}, 3rd edition, Wiley series in
  probability and mathematical statistics, pages 19--33. Wiley, New York,
  Chichester.
\newblock ISBN 0~471~40300.

\bibitem[{Huang et~al.(2022)Huang, Feng, Geng, and Qin}]{huang2022unifying}
Yichong Huang, Xiaocheng Feng, Xinwei Geng, and Bing Qin. 2022.
\newblock Unifying the convergences in multilingual neural machine translation.
\newblock In \emph{Proceedings of the 2022 Conference on Empirical Methods in
  Natural Language Processing}, pages 6822--6835.

\bibitem[{Jespersen(1922)}]{Jespersen:1922}
Otto Jespersen. 1922.
\newblock \emph{Language: Its Nature, Development, and Origin}.
\newblock Allen and Unwin.

\bibitem[{Kocmi et~al.(2022)Kocmi, Bawden, Bojar, Dvorkovich, Federmann,
  Fishel, Gowda, Graham, Grundkiewicz, Haddow et~al.}]{kocmi2022findings}
Tom Kocmi, Rachel Bawden, Ond{\v{r}}ej Bojar, Anton Dvorkovich, Christian
  Federmann, Mark Fishel, Thamme Gowda, Yvette Graham, Roman Grundkiewicz,
  Barry Haddow, et~al. 2022.
\newblock Findings of the 2022 conference on machine translation (wmt22).
\newblock In \emph{Proceedings of the Seventh Conference on Machine Translation
  (WMT)}, pages 1--45.

\bibitem[{Lester et~al.(2021)Lester, Al-Rfou, and Constant}]{lester2021power}
Brian Lester, Rami Al-Rfou, and Noah Constant. 2021.
\newblock The power of scale for parameter-efficient prompt tuning.
\newblock \emph{arXiv preprint arXiv:2104.08691}.

\bibitem[{Lewis et~al.(2019)Lewis, Liu, Goyal, Ghazvininejad, Mohamed, Levy,
  Stoyanov, and Zettlemoyer}]{lewis2019bart}
Mike Lewis, Yinhan Liu, Naman Goyal, Marjan Ghazvininejad, Abdelrahman Mohamed,
  Omer Levy, Ves Stoyanov, and Luke Zettlemoyer. 2019.
\newblock Bart: Denoising sequence-to-sequence pre-training for natural
  language generation, translation, and comprehension.
\newblock \emph{arXiv preprint arXiv:1910.13461}.

\bibitem[{Li et~al.(2023)Li, Zhou, Huang, Chen, and Chen}]{li2023eliciting}
Jiahuan Li, Hao Zhou, Shujian Huang, Shanbo Chen, and Jiajun Chen. 2023.
\newblock Eliciting the translation ability of large language models via
  multilingual finetuning with translation instructions.
\newblock \emph{arXiv preprint arXiv:2305.15083}.

\bibitem[{Lin et~al.(2022)Lin, Mihaylov, Artetxe, Wang, Chen, Simig, Ott,
  Goyal, Bhosale, Du et~al.}]{lin2022few}
Xi~Victoria Lin, Todor Mihaylov, Mikel Artetxe, Tianlu Wang, Shuohui Chen,
  Daniel Simig, Myle Ott, Naman Goyal, Shruti Bhosale, Jingfei Du, et~al. 2022.
\newblock Few-shot learning with multilingual generative language models.
\newblock In \emph{Proceedings of the 2022 Conference on Empirical Methods in
  Natural Language Processing}, pages 9019--9052.

\bibitem[{Liu et~al.(2022)Liu, Tam, Muqeeth, Mohta, Huang, Bansal, and
  Raffel}]{liu2022few}
Haokun Liu, Derek Tam, Mohammed Muqeeth, Jay Mohta, Tenghao Huang, Mohit
  Bansal, and Colin~A Raffel. 2022.
\newblock Few-shot parameter-efficient fine-tuning is better and cheaper than
  in-context learning.
\newblock \emph{Advances in Neural Information Processing Systems},
  35:1950--1965.

\bibitem[{Liu et~al.(2021)Liu, Ji, Fu, Tam, Du, Yang, and Tang}]{liu2021p}
Xiao Liu, Kaixuan Ji, Yicheng Fu, Weng~Lam Tam, Zhengxiao Du, Zhilin Yang, and
  Jie Tang. 2021.
\newblock P-tuning v2: Prompt tuning can be comparable to fine-tuning
  universally across scales and tasks.
\newblock \emph{arXiv preprint arXiv:2110.07602}.

\bibitem[{Liu et~al.(2023)Liu, Zheng, Du, Ding, Qian, Yang, and
  Tang}]{liu2023gpt}
Xiao Liu, Yanan Zheng, Zhengxiao Du, Ming Ding, Yujie Qian, Zhilin Yang, and
  Jie Tang. 2023.
\newblock Gpt understands, too.
\newblock \emph{AI Open}.

\bibitem[{Muennighoff et~al.(2022)Muennighoff, Wang, Sutawika, Roberts,
  Biderman, Scao, Bari, Shen, Yong, Schoelkopf
  et~al.}]{muennighoff2022crosslingual}
Niklas Muennighoff, Thomas Wang, Lintang Sutawika, Adam Roberts, Stella
  Biderman, Teven~Le Scao, M~Saiful Bari, Sheng Shen, Zheng-Xin Yong, Hailey
  Schoelkopf, et~al. 2022.
\newblock Crosslingual generalization through multitask finetuning.
\newblock \emph{arXiv preprint arXiv:2211.01786}.

\bibitem[{Ouyang et~al.(2022)Ouyang, Wu, Jiang, Almeida, Wainwright, Mishkin,
  Zhang, Agarwal, Slama, Ray et~al.}]{ouyang2022training}
Long Ouyang, Jeffrey Wu, Xu~Jiang, Diogo Almeida, Carroll Wainwright, Pamela
  Mishkin, Chong Zhang, Sandhini Agarwal, Katarina Slama, Alex Ray, et~al.
  2022.
\newblock Training language models to follow instructions with human feedback.
\newblock \emph{Advances in Neural Information Processing Systems},
  35:27730--27744.

\bibitem[{Papineni et~al.(2002)Papineni, Roukos, Ward, and
  Zhu}]{papineni2002bleu}
Kishore Papineni, Salim Roukos, Todd Ward, and Wei-Jing Zhu. 2002.
\newblock Bleu: a method for automatic evaluation of machine translation.
\newblock In \emph{Proceedings of the 40th annual meeting of the Association
  for Computational Linguistics}, pages 311--318.

\bibitem[{Pires et~al.(2023)Pires, Schmidt, Liao, and
  Peitz}]{Pires_Schmidt_Liao_Peitz_2023}
TelmoPessoa Pires, RobinM. Schmidt, Yi-Hsiu Liao, and Stephan Peitz. 2023.
\newblock Learning language-specific layers for multilingual machine
  translation.

\bibitem[{Popovi{\'c}(2015)}]{popovic2015chrf}
Maja Popovi{\'c}. 2015.
\newblock chrf: character n-gram f-score for automatic mt evaluation.
\newblock In \emph{Proceedings of the tenth workshop on statistical machine
  translation}, pages 392--395.

\bibitem[{Popovi{\'c}(2017)}]{popovic2017chrf++}
Maja Popovi{\'c}. 2017.
\newblock chrf++: words helping character n-grams.
\newblock In \emph{Proceedings of the second conference on machine
  translation}, pages 612--618.

\bibitem[{Post(2018)}]{post-2018-call}
Matt Post. 2018.
\newblock \href {https://www.aclweb.org/anthology/W18-6319} {A call for clarity
  in reporting {BLEU} scores}.
\newblock In \emph{Proceedings of the Third Conference on Machine Translation:
  Research Papers}, pages 186--191, Belgium, Brussels. Association for
  Computational Linguistics.

\bibitem[{Radford et~al.(2018)Radford, Narasimhan, Salimans, Sutskever
  et~al.}]{radford2018improving}
Alec Radford, Karthik Narasimhan, Tim Salimans, Ilya Sutskever, et~al. 2018.
\newblock Improving language understanding by generative pre-training.

\bibitem[{Radford et~al.(2019)Radford, Wu, Child, Luan, Amodei, Sutskever
  et~al.}]{radford2019language}
Alec Radford, Jeffrey Wu, Rewon Child, David Luan, Dario Amodei, Ilya
  Sutskever, et~al. 2019.
\newblock Language models are unsupervised multitask learners.
\newblock \emph{OpenAI blog}, 1(8):9.

\bibitem[{Raffel et~al.(2020)Raffel, Shazeer, Roberts, Lee, Narang, Matena,
  Zhou, Li, and Liu}]{raffel2020exploring}
Colin Raffel, Noam Shazeer, Adam Roberts, Katherine Lee, Sharan Narang, Michael
  Matena, Yanqi Zhou, Wei Li, and Peter~J Liu. 2020.
\newblock Exploring the limits of transfer learning with a unified text-to-text
  transformer.
\newblock \emph{The Journal of Machine Learning Research}, 21(1):5485--5551.

\bibitem[{Scao et~al.(2022)Scao, Fan, Akiki, Pavlick, Ili{\'c}, Hesslow,
  Castagn{\'e}, Luccioni, Yvon, Gall{\'e} et~al.}]{scao2022bloom}
Teven~Le Scao, Angela Fan, Christopher Akiki, Ellie Pavlick, Suzana Ili{\'c},
  Daniel Hesslow, Roman Castagn{\'e}, Alexandra~Sasha Luccioni, Fran{\c{c}}ois
  Yvon, Matthias Gall{\'e}, et~al. 2022.
\newblock Bloom: A 176b-parameter open-access multilingual language model.
\newblock \emph{arXiv preprint arXiv:2211.05100}.

\bibitem[{Schioppa et~al.(2023)Schioppa, Garcia, and Firat}]{schioppa2023cross}
Andrea Schioppa, Xavier Garcia, and Orhan Firat. 2023.
\newblock Cross-lingual supervision improves large language models
  pre-training.
\newblock \emph{arXiv preprint arXiv:2305.11778}.

\bibitem[{Sellam et~al.(2020)Sellam, Das, and Parikh}]{sellam2020bleurt}
Thibault Sellam, Dipanjan Das, and Ankur~P Parikh. 2020.
\newblock Bleurt: Learning robust metrics for text generation.
\newblock \emph{arXiv preprint arXiv:2004.04696}.

\bibitem[{Shin et~al.(2020)Shin, Razeghi, Logan~IV, Wallace, and
  Singh}]{shin2020autoprompt}
Taylor Shin, Yasaman Razeghi, Robert~L Logan~IV, Eric Wallace, and Sameer
  Singh. 2020.
\newblock Autoprompt: Eliciting knowledge from language models with
  automatically generated prompts.
\newblock \emph{arXiv preprint arXiv:2010.15980}.

\bibitem[{Singer et~al.(1954--58)Singer, Holmyard, and Hall}]{singer-whole}
Charles~Joseph Singer, E.~J. Holmyard, and A.~R. Hall, editors. 1954--58.
\newblock \emph{A history of technology}.
\newblock Oxford University Press, London.
\newblock 5 vol.

\bibitem[{Strötgen and Gertz(2012)}]{Martin-90}
Jannik Strötgen and Michael Gertz. 2012.
\newblock Temporal tagging on different domains: Challenges, strategies, and
  gold standards.
\newblock In \emph{Proceedings of the Eight International Conference on
  Language Resources and Evaluation (LREC'12)}, pages 3746--3753, Istanbul,
  Turkey. European Language Resource Association (ELRA).

\bibitem[{Superman et~al.(2000)Superman, Batman, Catwoman, and
  Spiderman}]{Superman-Batman-Catwoman-Spiderman-00}
S.~Superman, B.~Batman, C.~Catwoman, and S.~Spiderman. 2000.
\newblock \emph{Superheroes experiences with books}, 20th edition.
\newblock The Phantom Editors Associates, Gotham City.

\bibitem[{Touvron et~al.(2023)Touvron, Lavril, Izacard, Martinet, Lachaux,
  Lacroix, Rozi{\`e}re, Goyal, Hambro, Azhar et~al.}]{touvron2023llama}
Hugo Touvron, Thibaut Lavril, Gautier Izacard, Xavier Martinet, Marie-Anne
  Lachaux, Timoth{\'e}e Lacroix, Baptiste Rozi{\`e}re, Naman Goyal, Eric
  Hambro, Faisal Azhar, et~al. 2023.
\newblock Llama: Open and efficient foundation language models.
\newblock \emph{arXiv preprint arXiv:2302.13971}.

\bibitem[{Vaswani et~al.(2017)Vaswani, Shazeer, Parmar, Uszkoreit, Jones,
  Gomez, Kaiser, and Polosukhin}]{vaswani2017attention}
Ashish Vaswani, Noam Shazeer, Niki Parmar, Jakob Uszkoreit, Llion Jones,
  Aidan~N Gomez, {\L}ukasz Kaiser, and Illia Polosukhin. 2017.
\newblock Attention is all you need.
\newblock \emph{Advances in neural information processing systems}, 30.

\bibitem[{Vilar et~al.(2022)Vilar, Freitag, Cherry, Luo, Ratnakar, and
  Foster}]{vilar2022prompting}
David Vilar, Markus Freitag, Colin Cherry, Jiaming Luo, Viresh Ratnakar, and
  George Foster. 2022.
\newblock Prompting palm for translation: Assessing strategies and performance.
\newblock \emph{arXiv preprint arXiv:2211.09102}.

\bibitem[{Wallace et~al.(2019)Wallace, Feng, Kandpal, Gardner, and
  Singh}]{wallace2019universal}
Eric Wallace, Shi Feng, Nikhil Kandpal, Matt Gardner, and Sameer Singh. 2019.
\newblock Universal adversarial triggers for attacking and analyzing nlp.
\newblock \emph{arXiv preprint arXiv:1908.07125}.

\bibitem[{Wang et~al.(2022)Wang, Roberts, Hesslow, Le~Scao, Chung, Beltagy,
  Launay, and Raffel}]{wang2022language}
Thomas Wang, Adam Roberts, Daniel Hesslow, Teven Le~Scao, Hyung~Won Chung,
  Iz~Beltagy, Julien Launay, and Colin Raffel. 2022.
\newblock What language model architecture and pretraining objective works best
  for zero-shot generalization?
\newblock In \emph{International Conference on Machine Learning}, pages
  22964--22984. PMLR.

\bibitem[{Wang et~al.(2023)Wang, Panda, Karlinsky, Feris, Sun, and
  Kim}]{wang2023multitask}
Zhen Wang, Rameswar Panda, Leonid Karlinsky, Rogerio Feris, Huan Sun, and Yoon
  Kim. 2023.
\newblock Multitask prompt tuning enables parameter-efficient transfer
  learning.
\newblock \emph{arXiv preprint arXiv:2303.02861}.

\bibitem[{Wei et~al.(2022)Wei, Wang, Schuurmans, Bosma, Xia, Chi, Le, Zhou
  et~al.}]{wei2022chain}
Jason Wei, Xuezhi Wang, Dale Schuurmans, Maarten Bosma, Fei Xia, Ed~Chi, Quoc~V
  Le, Denny Zhou, et~al. 2022.
\newblock Chain-of-thought prompting elicits reasoning in large language
  models.
\newblock \emph{Advances in Neural Information Processing Systems},
  35:24824--24837.

\bibitem[{Wu et~al.(2021)Wu, Cheng, Wang, and Li}]{wu2021language}
Liwei Wu, Shanbo Cheng, Mingxuan Wang, and Lei Li. 2021.
\newblock Language tags matter for zero-shot neural machine translation.
\newblock \emph{arXiv preprint arXiv:2106.07930}.

\bibitem[{Xue et~al.(2020)Xue, Constant, Roberts, Kale, Al-Rfou, Siddhant,
  Barua, and Raffel}]{xue2020mt5}
Linting Xue, Noah Constant, Adam Roberts, Mihir Kale, Rami Al-Rfou, Aditya
  Siddhant, Aditya Barua, and Colin Raffel. 2020.
\newblock mt5: A massively multilingual pre-trained text-to-text transformer.
\newblock \emph{arXiv preprint arXiv:2010.11934}.

\bibitem[{Zhang et~al.(2023)Zhang, Haddow, and Birch}]{zhang2023prompting}
Biao Zhang, Barry Haddow, and Alexandra Birch. 2023.
\newblock Prompting large language model for machine translation: A case study.
\newblock \emph{arXiv preprint arXiv:2301.07069}.

\bibitem[{Zhang et~al.(2020)Zhang, Williams, Titov, and
  Sennrich}]{zhang2020improving}
Biao Zhang, Philip Williams, Ivan Titov, and Rico Sennrich. 2020.
\newblock Improving massively multilingual neural machine translation and
  zero-shot translation.
\newblock \emph{arXiv preprint arXiv:2004.11867}.

\bibitem[{Zhang et~al.(2022)Zhang, Roller, Goyal, Artetxe, Chen, Chen, Dewan,
  Diab, Li, Lin et~al.}]{zhang2022opt}
Susan Zhang, Stephen Roller, Naman Goyal, Mikel Artetxe, Moya Chen, Shuohui
  Chen, Christopher Dewan, Mona Diab, Xian Li, Xi~Victoria Lin, et~al. 2022.
\newblock Opt: Open pre-trained transformer language models.
\newblock \emph{arXiv preprint arXiv:2205.01068}.

\end{thebibliography}

% \section{Language Resource References}
% \label{lr:ref}
% \bibliographystylelanguageresource{lrec-coling2024-natbib}
% \bibliographylanguageresource{languageresource}

\end{document}